\documentclass[sigconf]{acmart}

\AtBeginDocument{%
  \providecommand\BibTeX{{%
    \normalfont B\kern-0.5em{\scshape i\kern-0.25em b}\kern-0.8em\TeX}}}

\acmConference[Conference acronym 'XX]{Make sure to enter the correct
  conference title from your rights confirmation emai}{June 03--05,
  2018}{Woodstock, NY}
\renewcommand\footnotetextcopyrightpermission[1]{} 
\copyrightyear{2023} 
\acmYear{2023} 
\setcopyright{acmlicensed}\acmConference[ICMI '23]{INTERNATIONAL CONFERENCE ON MULTIMODAL INTERACTION}{October 9--13, 2023}{Paris, France}
\acmBooktitle{INTERNATIONAL CONFERENCE ON MULTIMODAL INTERACTION (ICMI '23), October 9--13, 2023, Paris, France}
\acmPrice{15.00}
\acmDOI{10.1145/3577190.3614106}
\acmISBN{979-8-4007-0055-2/23/10}

\acmSubmissionID{1194}

\begin{document}

\title{Make Your Brief Stroke Real and Stereoscopic: 3D-Aware
Simplified Sketch to Portrait Generation}


\author{Yasheng Sun}
\authornote{Equal contribution.}
\affiliation{%
  \institution{Tokyo Institute of Technology}
  \city{Tokyo}
  \country{Japan}}
\email{sun.y.aj@m.titech.ac.jp}

\author{Qianyi Wu}
\authornotemark[1]
\affiliation{%
  \institution{Monash University}
  \city{Melbourne}
  \country{Australia}}
\email{qianyi.wu@monash.edu}

\author{Hang Zhou}
\authornotemark[1]
\affiliation{%
  \institution{Baidu Inc.}
  \city{Shanghai}
  \country{China}}
\email{zhouhang09@baidu.com}

\author{Kaisiyuan Wang}
\affiliation{%
  \institution{The University of Sydney}
  \city{Sydney}
  \country{Australia}}
\email{kaisiyuan.wang@sydney.edu.au}

\author{Tianshu Hu}
\affiliation{%
  \institution{Baidu}
  \city{Shanghai}
  \country{China}}
\email{hutianshu007@163.com}

\author{Chen-Chieh Liao}
\affiliation{%
  \institution{Tokyo Institute of Technology}
  \city{Tokyo}
  \country{Japan}}
\email{liao.c.aa@m.titech.ac.jp}

\author{Shio Miyafuji}
\affiliation{%
  \institution{Tokyo Institute of Technology}
  \city{Tokyo}
  \country{Japan}}
\email{miyafuji.s.aa@m.titech.ac.jp}

\author{Ziwei Liu}
\affiliation{%
  \institution{Nanyang Technological University}
  \city{Singapore}
  \country{Singapore}}
\email{ziwei.liu@ntu.edu.sg}

\author{Hideki Koike}
\affiliation{%
  \institution{Tokyo Institute of Technology}
  \city{Tokyo}
  \country{Japan}}
\email{koike@c.titech.ac.jp}

\renewcommand{\shortauthors}{Yasheng Sun, Hang Zhou, et al.}

\begin{abstract}
Creating the photo-realistic version of people's  sketched portraits is useful to various entertainment purposes. Existing studies only generate  portraits in the 2D plane with fixed views, making the results less vivid. In this paper, we present Stereoscopic Simplified Sketch-to-Portrait (SSSP), which explores the possibility of creating Stereoscopic 3D-aware portraits from simple contour sketches by involving 3D generative models. Our key insight is to design sketch-aware constraints
that can fully exploit the prior knowledge of a tri-plane-based 3D-aware generative model. Specifically, our designed region-aware volume rendering strategy and global consistency constraint further enhance detail correspondences during sketch encoding. Moreover, in order to facilitate the usage of layman users, we propose a Contour-to-Sketch module with vector quantized representations, so that easily drawn contours can directly guide the generation of 3D portraits. Extensive comparisons show that our method generates high-quality results that match the sketch. 
Our usability study verifies that our system is preferred by users.
\end{abstract}


\begin{teaserfigure}
  \centering
  \includegraphics[width=0.8\textwidth]{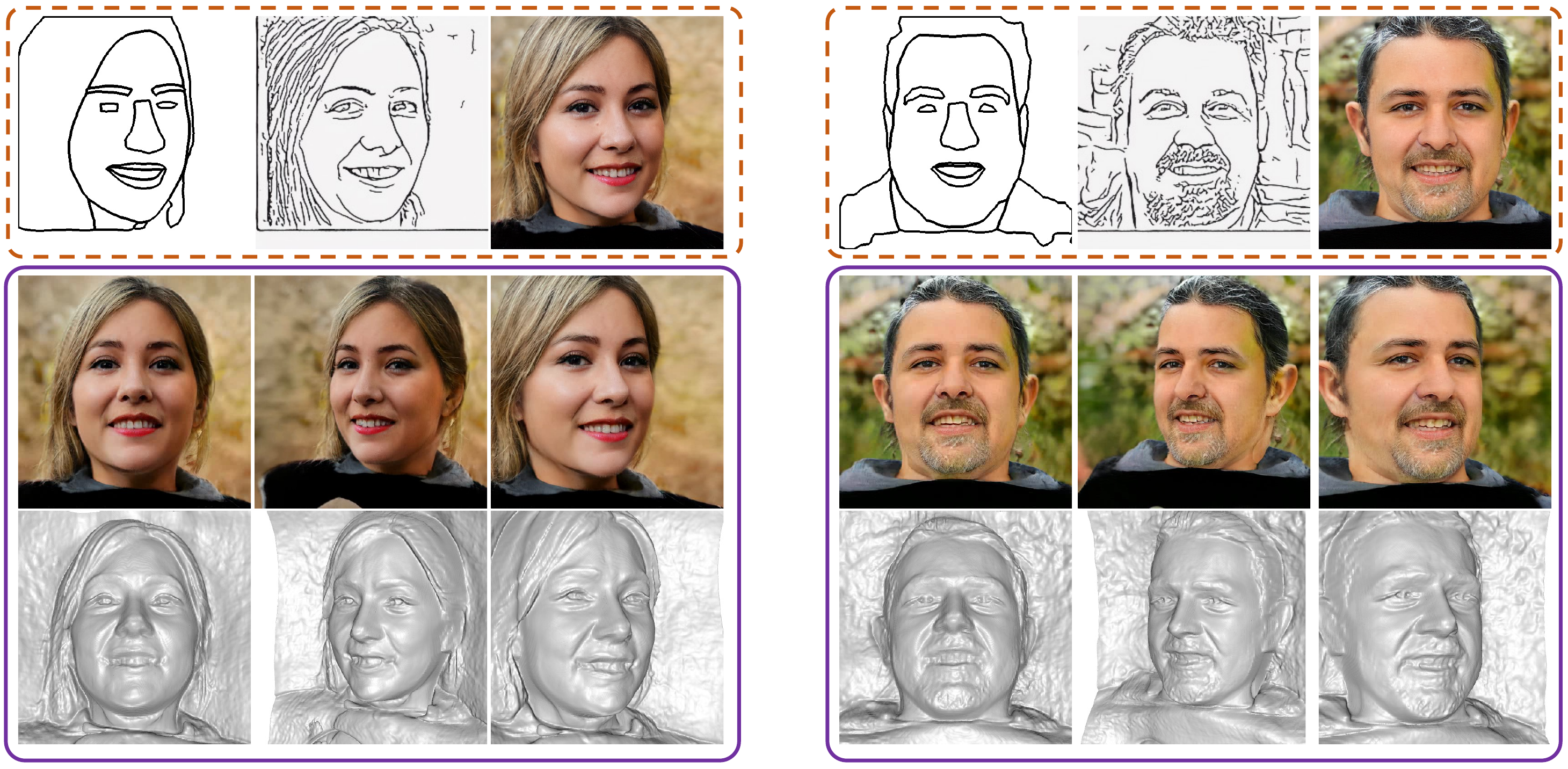}
  \caption{The visualization of our Stereoscopic Simplified Sketch-to-Portrait system. Given a simple contour, our system firstly convert it to a delicate sketch. Realistic portraits with consistent 3D geometry can be rendered by our system that aligned perfectly with the sketch. The contours and the sketches can both be modified by users for editable interactions.}
  \label{fig:teaser}
\end{teaserfigure}



\maketitle

\section{Introduction}

Sketching is a wonderful activity that can be enjoyed by people of all ages, which can be a valuable tool for both educational and creative purposes. Sketching faces, in particular, can be a means of expressing desired characters or identifying individuals.
However, there is often a significant gap between the casual sketches that many people create and realistic portraits. Bridging this gap is a valuable skill that can have many practical applications, such as entertainment through the use of augmented reality (AR) and virtual character design and creation for Metaverse.

With the development of generative models~\cite{goodfellow2014generative,karras2017progressive,karras2020analyzing} and their applications in image translation~\cite{pix2pix2017,CycleGAN2017,wang2018high}, semantic maps can already be faithfully transformed to real faces~\cite{lee2019maskgan,park2019semantic}. However, facial details such as wrinkles and hairstyles are lost in this kind of representation. As a result, researchers study specifically the problem of sketch-to-portrait generation~\cite{chen2008sketching,chen2009sketch2photo,dekel2018sparse,li2019linestofacephoto,chen2020deepfacedrawing,chen2021deepfaceediting}. They leverage real faces and their corresponding edge maps to build paired data and formulate the problem in a similar way as image translation tasks. 
Recent studies can generate realistic results aligned with various kinds of sketches~\cite{li2020deepfacepencil,chen2020deepfacedrawing}.
Nevertheless, previous methods are basically 2D-based, which means that their generated faces are constrained to the exact views as the sketches are drawn.
For scenarios such as character appearance design, a fixed-view image cannot be served as a sufficient reference. Even though face rotation techniques can be applied as post-processing, the results cannot guarantee 3D consistency. The stereoscopic nature of human faces makes it essential for designers to watch multi-view avatars.  Moreover, the next step towards creating vivid agents in virtual realities is to build a geometry-aware portrait in the 3D space. 

In this paper, we study the problem of creating stereoscopic portraits with high-quality 3D geometry from simple sketches. With the recent development of 3D-aware generative models~\cite{chan2021pi,chan2022efficient,or2022stylesdf,gu2021stylenerf}, researchers have succeeded in generating  realistic portraits by involving 3D volume rendering techniques~\cite{mildenhall2020nerf}. However, it is still challenging to build the connection between sketches and these models: 1) most 3D-aware GANs are generative models which do not support conditional inputs. The model building and training protocol designs are not trivial. 2) Unlike face parsing maps, sketches appear randomly without fixed semantics. It is difficult to constrain the generation process. 3) This task has rarely been explored before, leaving learning protocols uncertain.

To tackle this problem, we propose a system called \textbf{Stereoscopic Simplified Sketch-to-Portrait (SSSP)}, which renders 3D-consistent realistic portraits that are well aligned with simple sketches. The key is to \emph{delicately design sketch-aware constraints that can fully take the advantage of a tri-plane-based 3D-aware generative model.}   In detail, we propose to encode sketches into the prior latent space of tri-plane-based 3D-aware GAN~\cite{chan2022efficient}. A region-aware volume rendering strategy is proposed so that crucial regions can be directly rendered at higher resolutions for sketch matching. At the same time, we enforce symmetric sketches to produce symmetric 3D spaces, which greatly enhances global consistency.


Moreover, we ensure that our system is interactive-friendly. While previous steps build the mapping between a detailed sketch and real faces, it is non-plausible for amateur users to draw complicated sketches that match the sketch dataset. On the other hand, we assume that users should be able to sketch their desired portrait at both the coarse and fine levels. Thus a Contour-to-Sketch module is proposed to reduce the difficulty for the amateur user to use our system. This module is novelly designed based on vector quantized representations~\cite{esser2021taming} so that it can handle robust types of contour inputs. 


Our contributions can be summarized as follows: 
\begin{itemize}
    \item We propose to generate 3D-aware portraits from sketches by latent space encoding in tri-plane-based generative models with sketch-aware rendering constraints. 
    \item We design a novel Contour-to-Sketch module which can robustly convert simple contours to delicate sketches with vector quantized representations.
    \item  Extensive quantitative and qualitative experiments illustrate the effectiveness of our Stereoscopic Simplified Sketch-to-Portrait (SSSP) system. Studies on our developed interface proved that stereoscopic portraits are crucial to the satisfaction of users.
\end{itemize}
\section{Related Work}

\subsection{Sketch-based Portrait Synthesis} Synthesizing realistic portraits from given hand-drawn sketches has been a longstanding topic in both computer-human interaction and computer graphics communities. It is valuable for various applications in virtual reality, augmented reality and digital human creation. Early methods~\cite{chen2009sketch2photo, eitz2011photosketcher} retrieve local image patches from a large-scale human face collection and then compose these local patches back to an entire image according to the input sketches. However, these methods were designed without considering hallucination consistency, which inevitably leads to unrealistic results.
Taking advantage of deep neural networks, plenty of works~\cite{chen2018sketchygan, li2019linestofacephoto, chen2020deepfacedrawing, li2020deepfacepencil} are proposed recently for high-fidelity portrait synthesis by carefully devising the network architecture.
%
%
To improve user experience, DeepFaceDrawing~\cite{chen2020deepfacedrawing} introduces a real-time interactive system that enables users to input hand-drawn sketches with merely facial structures (\textit{i.e.}, five fixed components). 
%
%
However, almost all previous methods are conducted in the 2D space. Their results are static thus lack realism. Moreover, some of the methods are trained on a specific sketch domain. It will be difficult for them to handle rough or incomplete sketches.

Different from the previous approaches, we propose to generate 3D-aware realistic portraits with substantial 3D representations. Moreover, our proposed system employs a two-stage synthesis strategy for robust and high-quality portrait synthesis. 

\subsection{Generative 3D-aware Image Synthesis} 

%
Recent generative models~\cite{chan2021pi, schwarz2020graf, niemeyer2021giraffe, deng2022gram, cai2022pix2nerf, chen2022sem2nerf, chan2022efficient, sun2022ide} involving neural implicit representation (INR) techniques~\cite{mildenhall2020nerf} have demonstrated great potential on high-quality image synthesis.
Specifically, EG3d~\cite{chan2022efficient} proposed an efficient tri-plane hybrid explicit-implicit 3D representation to synthesize high-resolution images in real-time with view consistency together with high-quality 3D geometry.
%
Despite promising image quality, these approaches cannot provide interactive local editing on the synthesized portraits.

Some concurrent works~\cite{sun2022fenerf, chen2022sem2nerf, sun2022ide} have made their attempts on conditional generation setting. 
FENeRF~\cite{sun2022fenerf} proposed to use the semantic mask for editing the 3D facial volume of the target portrait via GAN inversion~\cite{karras2020analyzing}.
However, it fails to support high-resolution image synthesis and real-time user-interactive applications due to its optimization-based inversion.
%
%
In order to achieve real-time interactive editing, IDE-3D~\cite{sun2022ide} introduced a high-resolution semantic-aware 3D generative model, which enables disentangled control over local shape and texture by leveraging a hybrid GAN inversion strategy.
Although IDE-3D is exploring similar settings (\textit{i.e.}, semantic-aware 3D portrait synthesis and editing), our approach mainly focuses on building a user-friendly real-time interactive system even for non-artists by taking only hand-drawn sketches as input, which are much rougher and more abstract signals than the semantic mask used in IDE-3D.
\section{Methodology}
\begin{figure*}
\begin{center}
\includegraphics[width=0.8\linewidth]{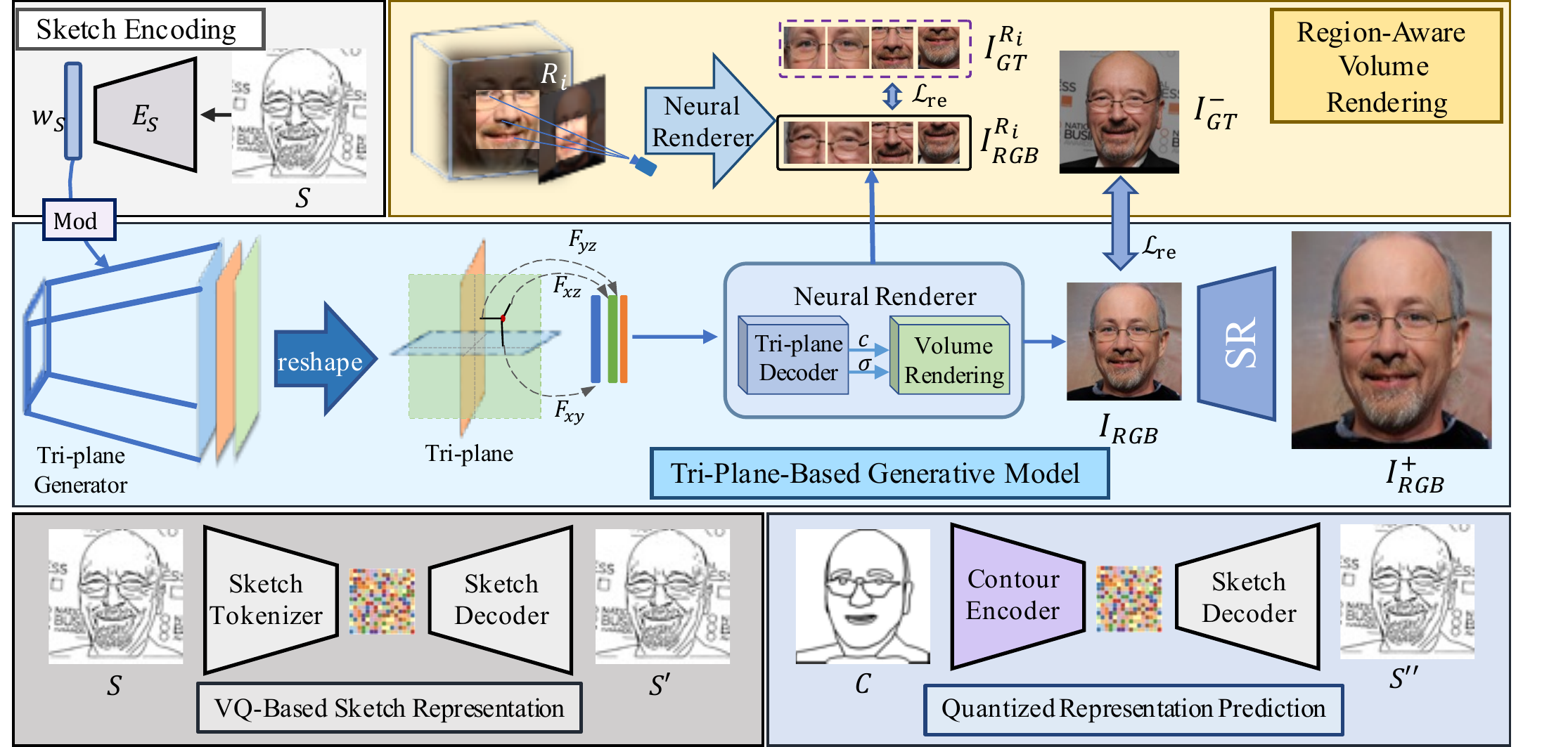}
\end{center}
\caption{\textbf{Pipeline of our proposed Stereoscopic Simplified Sketch-to-Portrait (SSSP)}. The overall framework consists of two major parts. The first part is the \textbf{Contour-to-Sketch Module}. At first, the user will edit a simplified facial contour and feed it to this module to obtain a predicted detailed sketch via a vector-quantized (VQ) delicate sketch representation. Another part is the \textbf{Sketch Guided 3D-Aware Portrait Generation module}, which aims to transfer the predicted delicate sketch to a tri-plane feature and render it as the final high-fidelity portrait image. } 
\label{fig:pipeline}
\vspace{-10pt}
\end{figure*}

In this section, we dive into the details of our proposed system, \textbf{Stereoscopic Simplified Sketch-to-Portrait (SSSP)}. Our target is to create 3D-aware realistic human portraits from 2D sketches with simple user interaction. 

\subsection{3D-Aware Portrait Generation from Sketch}\label{sec:3.1}
%
\noindent\textbf{Problem Formulation.} Given an arbitrary sketch $S$, our training goal is to recover the implicit or explicit 3D representation of it, and render it back to the real domain following the guidance of its paired image $I_{GT}$. To
 fulfill high-fidelity 3D-awareness,
we take inspiration from the recent Nerf based 3D generative models~\cite{gu2021stylenerf,chan2022efficient,or2022stylesdf} that are able to produce high quality portraits via volume rendering in the 3D space. 
%
%
%
%
Our solution is to pose this challenging task in the form of ``3D GAN inversion'', where an encoder is trained to learn the prior latent space of a pre-trained generator. Thus the task is formulated into the problem of render-guided encoder learning.

\noindent\textbf{Preliminaries on Tri-Plane-Based Generator.}
The tri-plane-based generator~\cite{chan2022efficient} that enjoys high 3D consistency and rendering quality, is one of the best choices.
As shown in the blue box of Fig.~\ref{fig:pipeline}, it leverages an explicit tri-plane 3D representation which stores 3D scene information into three orthogonal feature images of size $2h \times 2h \times C$. For each point $p\in\mathbb{R}^3$ in the 3D space, its projections on the three planes ($xy/xz/yz$-plane) query corresponding interpolated features $(F_{xy}, F_{xz}, F_{yz})$ and aggregate them to $F$ for representing the information of this spatial location. Afterwards, a tiny decoder processes $F$ to predict the color feature and density of this position $p$. These tri-plane feature images can be effectively sampled from a latent  vector $z$ with a StyleGAN2 generator~\cite{karras2020analyzing}. 

Volume rendering~\cite{max1995optical} is then performed by casting rays from the camera as NeRF~\cite{mildenhall2020nerf}. Differently, here the color features are accumulated to a 2D  feature image $I_F$ also with $C$ channels. 
In the original design, a moderate resolution of $h \times h$ is selected. $I_F$ is then sent into two different paths. One is to directly produce an RGB image $I_{RGB}$, and the other is to upsample its spatial resolution through more layers of modulated convolutions~\cite{karras2020analyzing}. Thus a high-resolution RGB image $I^+_{RGB}$ can be rendered. 


\noindent\textbf{Basic Learning Objective.}
Our next goal is to accurately restore the radiance fields (tri-planes) that render images that strictly match the input sketch. 
As the tri-planes are created by the StyleGAN2 generator, the sketches could be intuitively encoded to the $W$ or $W^+$ space. 
The encoder $E_S$ encodes the sketch into a feature $w_S$ in the $W$ space and sent into the generator. While two images $I_{RGB}$ and $I^+_{RGB}$ are produced within the generator, we identify that the directly rendered low-resolution image $I_{RGB}$ is more suitable for applying constraints, as the backward path is shorter. The first training constraint is to recover the downsampled ground truth image $I^-_{GT}$. The basic training objectives are the L1 reconstruction loss and the VGG perceptual loss~\cite{wang2018high,park2019semantic} between the rendered results and ground truth,
\begin{align}
\label{eq:1}
    \mathcal{L}_{re}(I^-_{GT}, I_{RGB}) = &\| I^-_{GT} - I_{RGB} \|_1 +  \nonumber \\ 
    & \sum_{m=1}^{N_{vgg}}\|\text{VGG}_m({I}^{-}_{GT}) - {\text{VGG}}_{m}({I}_{RGB}) \|_1,
\end{align}\label{eqn:img_rec}
where $\text{VGG}_{m}$ denotes the $m$-th layer's results of a VGG19 network. We have also tried computing losses on the final outputs and found similar results.


\subsection{Sketch-Aware Rendering Constraints}
However, using the universal pixel-level constraint cannot guarantee the detailed consistency between drawn lines and textures. While the original tri-plane generator is trained without local perception, the entire tri-plane formulation is vulnerable to the tiny changes in the $W$ space of the generator. To tackle this problem, we propose two novel sketch-aware rendering constraints, so that the model can grasp the local information from input sketch and achieve global consistency.

\subsubsection{Region-Aware Volume Rendering Strategy for Encoding Sketches} 

While depicting a human portrait, some local areas such as the eyes, nose and mouse play important roles. In practice, users usually spend more strokes on these regional parts.
Thus we argue that \emph{the network should be sensitive to local sketch differences}. 


To this end, a region-aware volume rendering  strategy is proposed to \emph{encourage the mapped latent code to focus on important local sketch areas}. 
%
Specifically, 4 particular regions $\{R_i | i \in [1, 4]\}$ representing the left and right eyes, the nose and the mouth are selected on our rendered low-resolution image ${I}_{RGB}$ of size $h\times h$.
The corresponding size of the $i$-th region is $\{\delta_i h \times \delta_i h\}$.
We perform volume rendering on each of these regions by densely sampling more rays, and then synthesize RGB images $I^{R_i}_{RGB}$ which has the same resolution as $I_{RGB}$. The reconstruction constraints are applied to all the rendered local regions. The loss can be written as:
\begin{align}
\label{eq:2}
    \mathcal{L}^R_{re} = \sum_{i=1}^{4}\mathcal{L}_{re}(I^{R_i}_{GT}, I^{R_i}_{RGB}),
\end{align}
where $I^{R_i}_{GT}$ is the cropped and resized region of $R_i$.
Compared with $R_i$ on $I_{RGB}$,  $I^{R_i}_{RGB}$ preserves much more details. With the additional refinement of these important patches, the encoder can learn the fine-grained correlations between sketches and portraits, leading to more convincing results.



\subsubsection{Symmetric Constraint} 
%
\begin{figure}
\begin{center}
\includegraphics[width=1.0\linewidth]{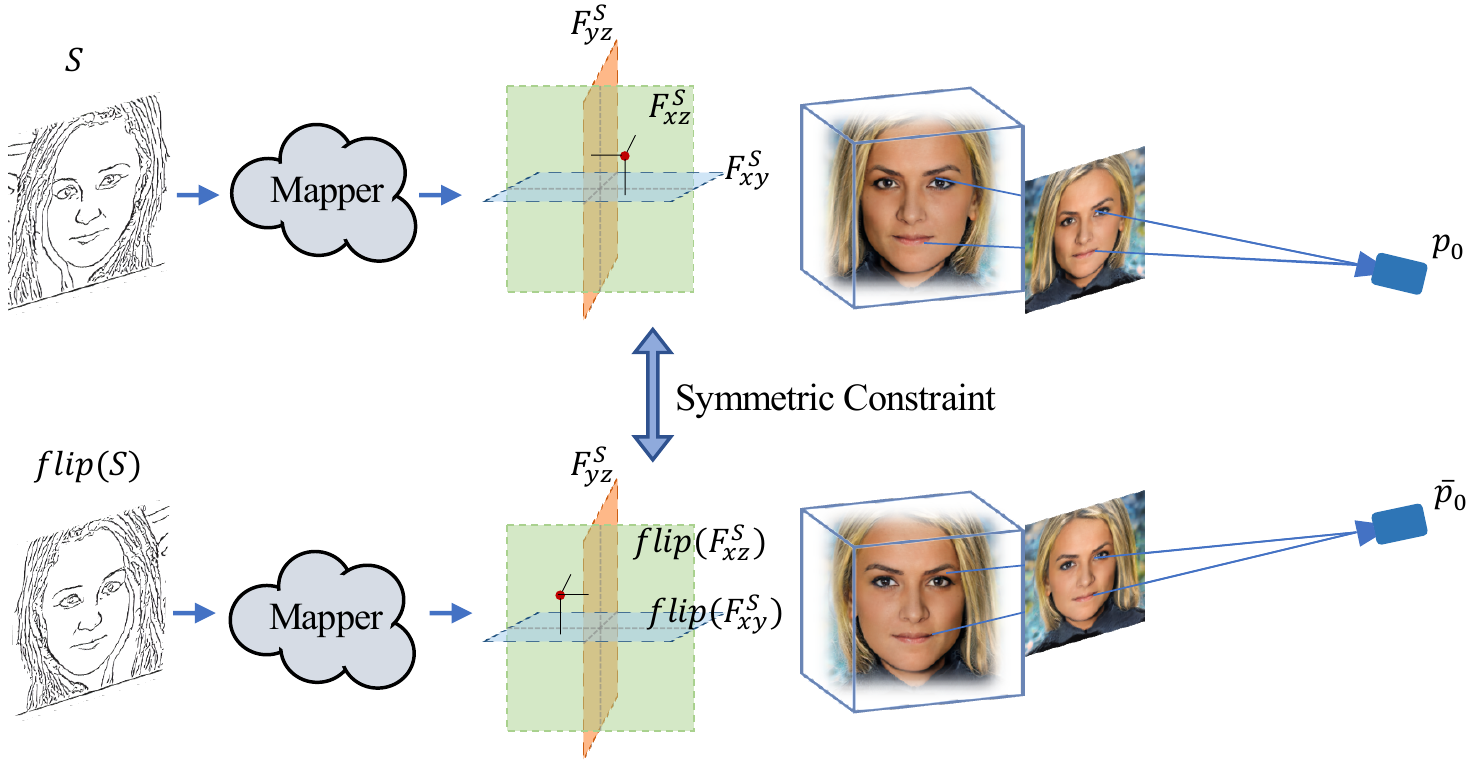}
\end{center}
\caption{\textbf{Symmetric Constraint}. We design a tri-plane based mirror constraint to further enforce the global consistency of input sketch and output image. As shown above, the mapped tri-plane feature of the flip sketch should maintain consistency for the symmetric sampling points about $yz$-plane.} 
\vspace{-10pt}
\label{fig:tmc}
\end{figure}

Involving the region-aware volume rendering strategy could significantly boost the model's sketch consistency on local details. However, it is also challenging to capture the global information such as the facial outline. As demonstrated in \cite{chan2022efficient}, certain global attributes like general facial expressions are biased towards the input camera pose. 
This phenomenon still exists when the model is trained with camera pose serving as conditions. An underlining cause is that the constraints are too limited to force the encoder network to capture the global pattern.  


With explicit 3D information leveraged in our setting, global constraints can be more easily added to both the 3D representations and the rendered image.
Particularly, the simple flipping data augmentation is used here. We would expect the flipped sketch representing a person with symmetric geometry in the 3D space. 

In detail,
we take inspiration from unsupervised symmetric learning of 3D  faces~\cite{wu2020unsupervised}, but explore a new solution on tri-plane. 
Without loss of generality, we locate the portrait in the origin of tri-plane space and look towards the $y$-axis without any rotation.
Our insight is that the flipping of an image in the 2D space leads to a flip along the $yz$-plane of its tri-plane space. 
 More specifically, if one camera is located at position $p_0$ in the tri-plane space and takes a picture of the portrait, we can move the camera to the symmetric position $\bar{p_0}$  about $yz$-plane and flip the tri-plane features of the $xy$ and  $xz$-plane to get a horizontally flipped image. 
 We denote $S$ as an input sketch  and $\bar{S}$ as its horizontal flipped sketch. By encoding them to the generator, two tri-planes, $F^S(p, p_0)=(F^S_{xy}, F^S_{xz}, F^S_{yz})$ and $F^{\bar{S}}(p', \bar{p_0})=(F^{\bar{S}}_{xy}, F^{\bar{S}}_{xz}, F^{\bar{S}}_{yz})$, can be generated, respectively. Here $p$ and $p'$ are symmetric sample points about the $yz$-plane, $p_0$ and $\bar{p_0}$ are symmetric camera pose about $yz$-plane as shown in Fig.~\ref{fig:tmc}. As the mirror constraint of 3D face naturally existed, we could further encourage the tri-plane features to satisfy the following equations:
\begin{equation}
\left\{
\begin{array}{lr}
    F^S_{xy} = \textbf{flip}(F^{\bar{S}}_{xy}, x), &\\
    F^S_{xz} = \textbf{flip}(F^{\bar{S}}_{xz}, x), & \\
    F^{S}_{yz} = F^{\bar{S}}_{yz} &,
\end{array}
\right.
\end{equation}
where $\textbf{filp}(F, x)$ means flipping the coordinates of tri-plane F along the $x$-axis. Therefore, we introduce a novel constraint, which is termed as \textbf{Symmetric Constraint} on the tri-plane feature together with the image reconstruction loss in Eq.~\ref{eq:1} and Eq.~\ref{eq:2}.

\subsection{Coutour-to-Sketch Generation via Vector-Quantized Representation}~\label{sec:3.2}
It is very challenging for amateur users to draw a detailed sketch as the input data we use in the portrait generation module (\ref{sec:3.1}). Also, painters usually produce contours first before adding details. Thus it seems more reasonable and convenient that users could choose to produce and edit both the coarse and fine levels of sketches. This leads to a new challenge: how to bridge the domain gap between a coarse contour and the detailed sketch input for our model?

To mitigate this issue, 
we propose the Contour-to-Sketch module to transfer simplified contour to a detailed sketch under an image translation protocol~\cite{pix2pix2017}. 
However, we find that directly applying existing models~\cite{wang2018high,park2019semantic} on this type of data will lead to failure cases. For example, some important parts of the face such as the  eyes will be missing.

Thus it is essential to seek a more robust design. We observe that the stroke types of facial sketches seem limited from a local perspective. Thus we propose to build a discrete codebook for the  detailed sketches via vector-quantized (VQ) representations~\cite{van2017neural}. This guarantees that all generated results lie in the distribution of the realistic sketches. The learning procedures are introduced in the following.

\noindent\textbf{Quantized Sketch Representation.} Firstly, we derive a learnable codebook with all data of detailed sketches.  A sketch tokenizer $T_S$ and a sketch decoder $D_S$ are designed following VQGAN~\cite{esser2021taming}. Given an arbitrary detailed sketch from the dataset, $S$, it is first encoded to a feature map $\hat{\textbf{z}}_k$ through the tokenizer. Then each value on its spatial vector is later transformed into discrete values according to the closest codebook entry. During decoding, these tokens can be de-quantized to feature vectors through querying the features stored in the codebook. They can be recovered back to a sketch with the decoder $D_S$.
The encoding-decoding scheme is simplified as:
\begin{align}
    S' = D_S(\mathbf{q^{-1}}(\mathbf{q}(T_S(S)))),
\end{align}
where $\mathbf{q}$ and $\mathbf{q^{-1}}$ denote the vector quantization and de-quantization operations.

\noindent\textbf{Quantized Representation Prediction.} Once the model is trained, our next step is to map the coarse contour to the learned entities in the codebook. We design the contour-to-token mapping in a teacher-student manner.
 Specifically, we aim to train a new encoder $E_C$ for the coarse contour $C$ and map it into the latent space of learnable codebook $\mathcal{V}$. The supervisions are two-folded. Firstly on each local patch position, the cross-entropy softmax classification loss is performed for predicting the correct token encoded by $T_S$. The next supervision is to directly constrain the distances between the encoded feature $E_C(C)$ from the contour and the de-quantized feature value $\mathbf{q^{-1}}(\mathbf{q}(T_S(S)))$. The final mapping procedure would be:
\begin{align}
    S'' = D_S(\mathbf{q^{-1}}(\mathbf{q}(E_C(C)))).
\end{align}

With the help of this module, rich details can be added to basic contour sketches robustly. This module is essential to the usability of our model. 






\section{Experiments}
\subsection{Experimental Settings}
\noindent{\textbf{Datasets.}} For training the Sketch-Guided Stereoscopic Portrait Generation
module, we leverage the dataset proposed in DeepFaceDrawing~\cite{chen2020deepfacedrawing}. It is created using the faces and parsing maps provided in CelebAMask-HQ~\cite{lee2019maskgan} and then processed with Photocopy and sketch simplification procedures. It contains 17K pairs of sketch-image pairs.
As for training the Contour-to-Sketch module, we process the above-mentioned dataset and extract contours using the tools provided in~\cite{chen2022sem2nerf}.

\noindent{\textbf{Implementation Details.}}
The model of our generator is the same as EG3D~\cite{chan2022efficient}. The resolution of the final output image $I^+_{RGB}$ is $512\times 512$. The size of the low-solution image is $128 \times 128$ ($h$ = 128). The StyleGAN2 generator backbone produces a feature map of size $256 \times 256 \times 96$, which is later reshaped to the tri-planes, each of size $256 \times 256 \times 32$. The sketch encoder architecture is designed as ResNet34~\cite{he2016deep}. Our Sketch Guided Portrait generation model is trained on 2 Tesla A100 GPUs for 3 days.

\noindent{\textbf{Quantitative Evaluation Metrics.}}
We conduct quantitative evaluations on metrics that have previously been used in the field of image generation and image quality assessment. We use \textbf{PSNR}, \textbf{SSIM}\cite{wang2004image}, \textbf{CPBD}~\cite{narvekar2009no} to measure the generated images over image fidelity, structure similarity, and sharpness perspective. We also report two metrics, \textbf{FID}~\cite{heusel2017gans} and \textbf{IS}~\cite{salimans2016improved}, commonly used in the image generation task to evaluate the image quality in latent distribution.


\noindent{\textbf{Comparison Methods.}} To the best of our knowledge, there does not exist a method that shares our setting. Thus we leverage several related state-of-the-art methods including pSp~\cite{richardson2021encoding}, pix2pixHD~\cite{wang2018high}, and DeepFaceDrawing~\cite{chen2020deepfacedrawing}.

Pix2pixHD is still one of the best general image-to-image translation frameworks, which supports various types of guidance. We directly train a sketch-to-portrait model using their released codes. pSp is proposed for StyleGAN2 inversion, it works in a similar way to ours and can be used for performing image translation while fixing the generator. DeepFaceDrawing opens an online interactive system, allowing users to input hand-drawn sketches.

\begin{figure}[t]
\begin{center}
\includegraphics[width=1.0\linewidth]{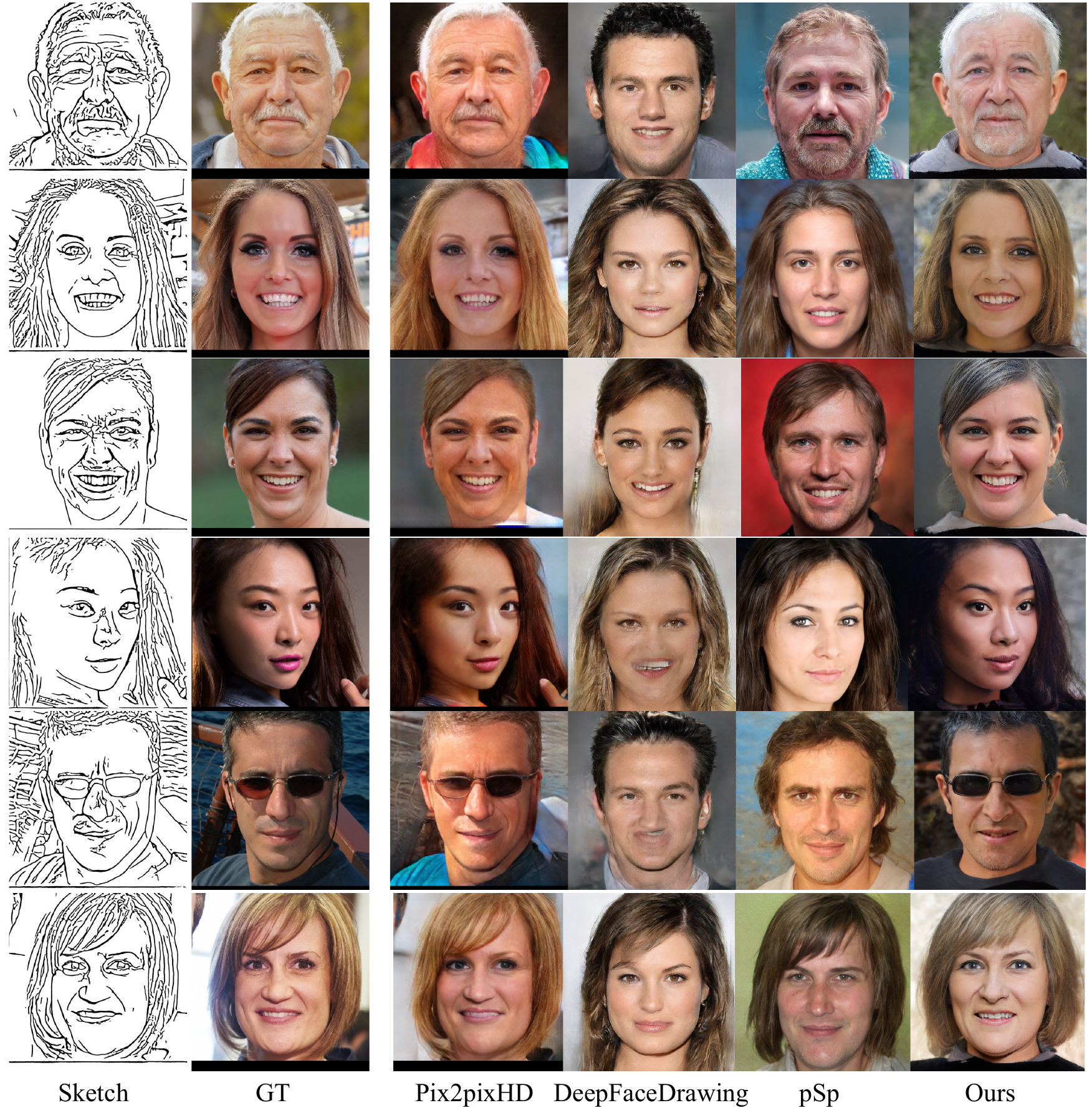}  
\end{center}
\vspace{-10pt}

\caption{\textbf{Qualitative Comparisons with Different Methods.} Our results achieve the best stroke consistency and are of high quality.
} 
\vspace{-15pt}
\label{fig:fix}
\end{figure}

\begin{figure*}
\begin{center}
\includegraphics[width=0.8\linewidth]{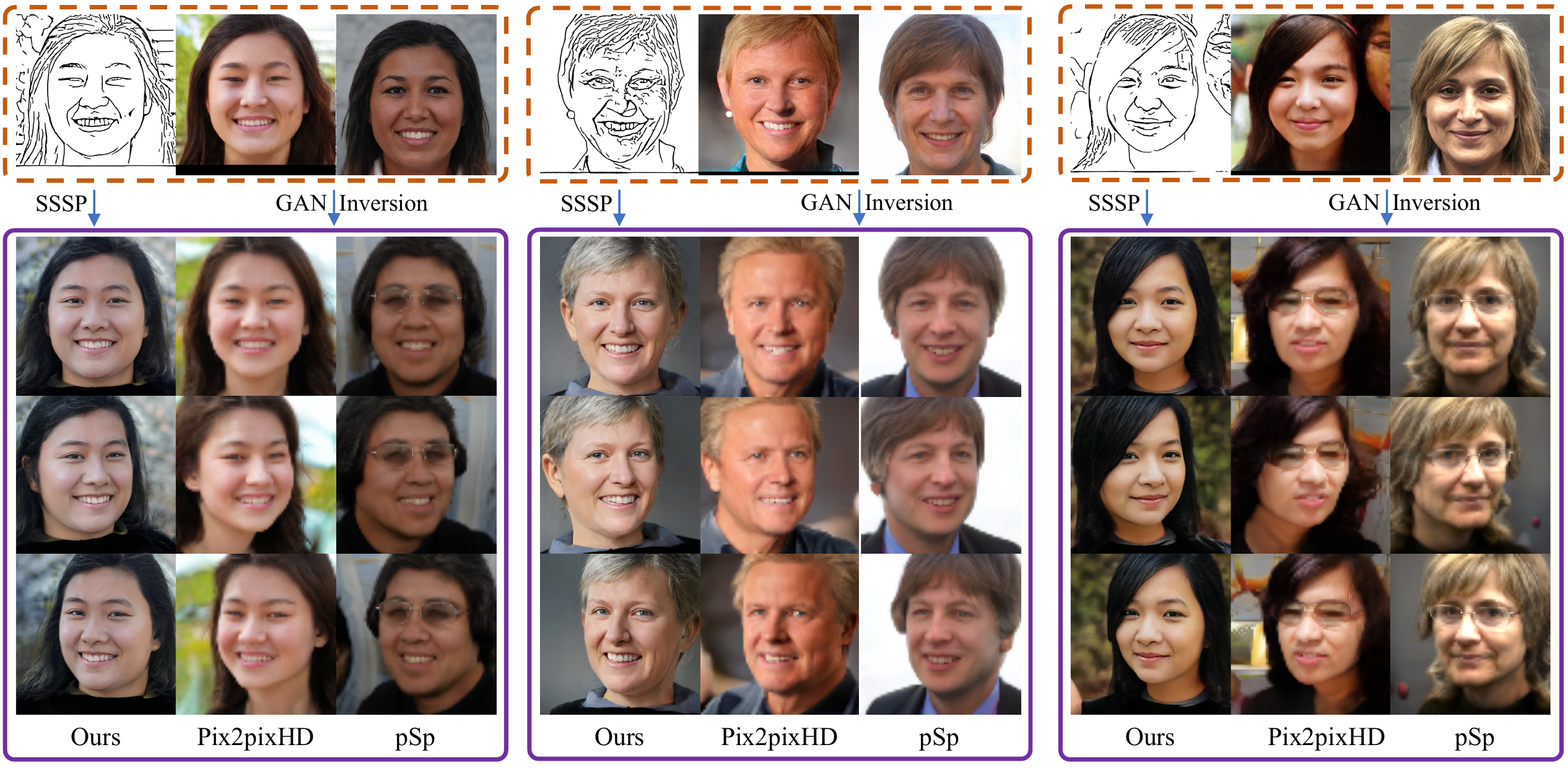}
\end{center}
\vspace{-10pt}
\caption{\textbf{Qualitative comparison on Free-View Synthesis}. We compare our SSSP with the state-of-the-art image generation methods pSp~\cite{richardson2021encoding} and pix2pixHD~\cite{wang2018high}. To better visualize the multi-view consistency of the generated views, we provide three views (\textit{i.e.}, front, left, and right) for the results from each method.
} 
\vspace{-5pt}
\label{fig:free}
\end{figure*}

\subsection{Evaluation on Fix-View Synthesis}
We first evaluate the generation quality of our method under a fix-view setting quantitatively and qualitatively. For fair comparisons, we use the sketches extracted from FFHQ~\cite{karras2019style} dataset as input without using the Contour-to-Sketch module. The qualitative results are shown in Fig.~\ref{fig:fix}. We can observe that DeepFaceDrawing can only synthesize results roughly consistent with the input sketches, which even lack necessary identity information and reasonable texture on some local facial components. pSp produces high-quality portraits for each input sketch, while its results suffer from facial structure or hairstyle mismatch issues and occasionally incorrect generated genders. Pix2pixHD generates the most satisfactory resultant portraits among all the comparison methods with consistent facial boundaries, semantics, and detailed textures. Our SSSP system achieves comparable performance with pix2pixHD, where the generated high-quality results match the input sketches well but with the identity slightly changed.

\setlength{\tabcolsep}{6pt}
\begin{table}[t] \footnotesize

\caption{Quantitative comparisons on Free-View Synthesis.} 
\label{tab:free}

\vspace{-5pt}

\begin{center}  
\begin{tabular}{lcccccc} 
\toprule
Method & $\text{SSIM}  \uparrow$ & \text{PSNR} $\uparrow $  & FID $\downarrow $ & IS $\uparrow $ & CPBD $\uparrow $ \\

\midrule
  \\ 
pSp (w inv)  & 0.363 &  10.618 & 96.19 & 2.86 & 0.591  \\ 
pix2pixHD (w inv) & 0.441 & 13.262 & 86.99 & \textbf{3.28} & 0.596 \\

\hline
SSSP ($W^{+}$ space) & \textbf{0.464} & 14.594 & \textbf{86.77} &  2.42 & 0.563 \\
SSSP (W space) & 0.448 & \textbf{15.237} & 98.50 & 2.96  & \textbf{0.625} \\
\bottomrule
\end{tabular}
\end{center}
\vspace{-15pt}
\end{table}

\subsection{Evaluation on Free-View Synthesis}
Since our goal is to render 3D-aware portraits from given sketches, we also design additional experiments under a free-view-point setting to evaluate the novel-view synthesis ability of our system. The novel view synthesis results of 2D StyleGAN2-based models~\cite{shen2020interfacegan} do not enjoy 3D consistency. Thus as common practice, the results generated by previous methods are converted also to the 3D space for novel view rendering. For fair comparisons, we directly leverage the model of our generator EG3D~\cite{chan2022efficient} for performing 3D GAN inversion. 

Specifically, the results of Pix2pixHD and pSp are firstly generated on the 2D plane. Then we map the image to a corresponding latent code in EG3D via optimization-based GAN-inversion. Based on the recovered tri-plane representation, free-view images can be directly obtained by changing the camera pose information.  Note that DeepFaceDrawing~\cite{chen2020deepfacedrawing} focuses on frontal face synthesis and re-training it on our training data with various poses will inevitably result in degradation in the generation quality, thus, we do not involve it in this comparison.

\noindent{\textbf{Quantitative Comparisons.}}
We first perform quantitative comparison between our SSSP and its counterparts. The results are reported in Table~\ref{tab:free} in which the two comparison methods are denoted as ``pix2pixHD (w inv)'' and ``pSp (w inv)'', respectively. Our model outperforms its counterparts on most of the evaluation metrics, which demonstrates the high-quality image generation ability of our approach.

\noindent{\textbf{Qualitative Comparisons.}}
Qualitative comparisons are depicted in Fig.~\ref{fig:free}. It can be seen that both pSp and pix2pixHD suffer from blurry textures under novel views. We analyze that most problems are caused by the imperfect 3D inversion of their results. Though we have set a long optimization time of about 20 minutes and tried altering the inversion weights, the inversion results of EG3D on these methods are still unsatisfactory. On the other hand, with our encoder, our method produces faces in less than a second. Moreover, the generation quality of our method is clearly better. This proves the effectiveness of our method.

\subsection{Ablation Study}
To further evaluate the contributions of our proposed techniques, we conduct an ablation study and the results are reported in Table~\ref{tab:ablation}. Concretely, we construct three variants. They are the model without region-aware rendering strategy (w/o RAS), the model without symmetric constraint (w/o SC), and encoding the $W^+$ space. Among all the variants,``w/o RAS'' achieves the worst score on CPBD, which demonstrates that our proposed RAS can effectively alleviate blurry textures in the results. ``w/o SC'' achieves the worst SSIM and PSNR performance when omitting this proposed constraint. This indicates that the symmetric constraint can effectively enhance the learning of the encoder. Furthermore, we also attempt to explore the generation ability of our model when the sketches are encoded to $W^{+}$ space instead of $W$ space in our full pipeline ``Full model (W space)''. Compared with our full model, ``$W^{+}$ space'' obtains comparable performance on the metrics related to image quality but achieves a much lower score on CPBD.  This indicates that encoding the sketches to W space contributes to better 3D portrait synthesis quality on novel views.

Despite the quantitative comparisons, we also provide visualisation results of these variants. Compared with our full model, ``w/o RAS'' tends to render over-smooth facial textures on local parts (\textit{e.g.}, eyes, mouth and teeth) which degrade the quality of generated images. While for ``w/o SC'', we provide an additional comparison in the last two rows at the bottom with a pair of symmetric input sketches. ``w/o SC'' fails to reconstruct symmetric images for these two sketches, which intuitively demonstrates the effectiveness of our proposed SC. By using all the components in our full model, we are able to synthesize high-quality and multi-view consistent portraits.

\setlength{\tabcolsep}{5pt}
\begin{table}[t] \footnotesize

\caption{Quantitative Results of Ablation Study.}
\label{tab:ablation}
\begin{center}  
\begin{tabular}{lcccccc} 
\toprule
Method & $\text{SSIM}  \uparrow$ & \text{PSNR} $\uparrow $  & FID $\downarrow $ & IS $\uparrow $ & CPBD $\uparrow $ \\

\midrule

w/o RAS & 0.420 & 14.128 & 101.24 & 2.43 & 0.471 \\
w/o SC & 0.412 & 13.662 & 102.53 & 2.69 & 0.564 \\
Full Model (W space) & \textbf{0.448} & \textbf{15.237} & \textbf{98.50} & \textbf{2.96}  & \textbf{0.625} \\

\bottomrule
\end{tabular}
\end{center}
\vspace{-15pt}
\end{table}


\begin{figure}[t]
\begin{center}
\vspace{-5pt}
\includegraphics[width=1.0\linewidth]{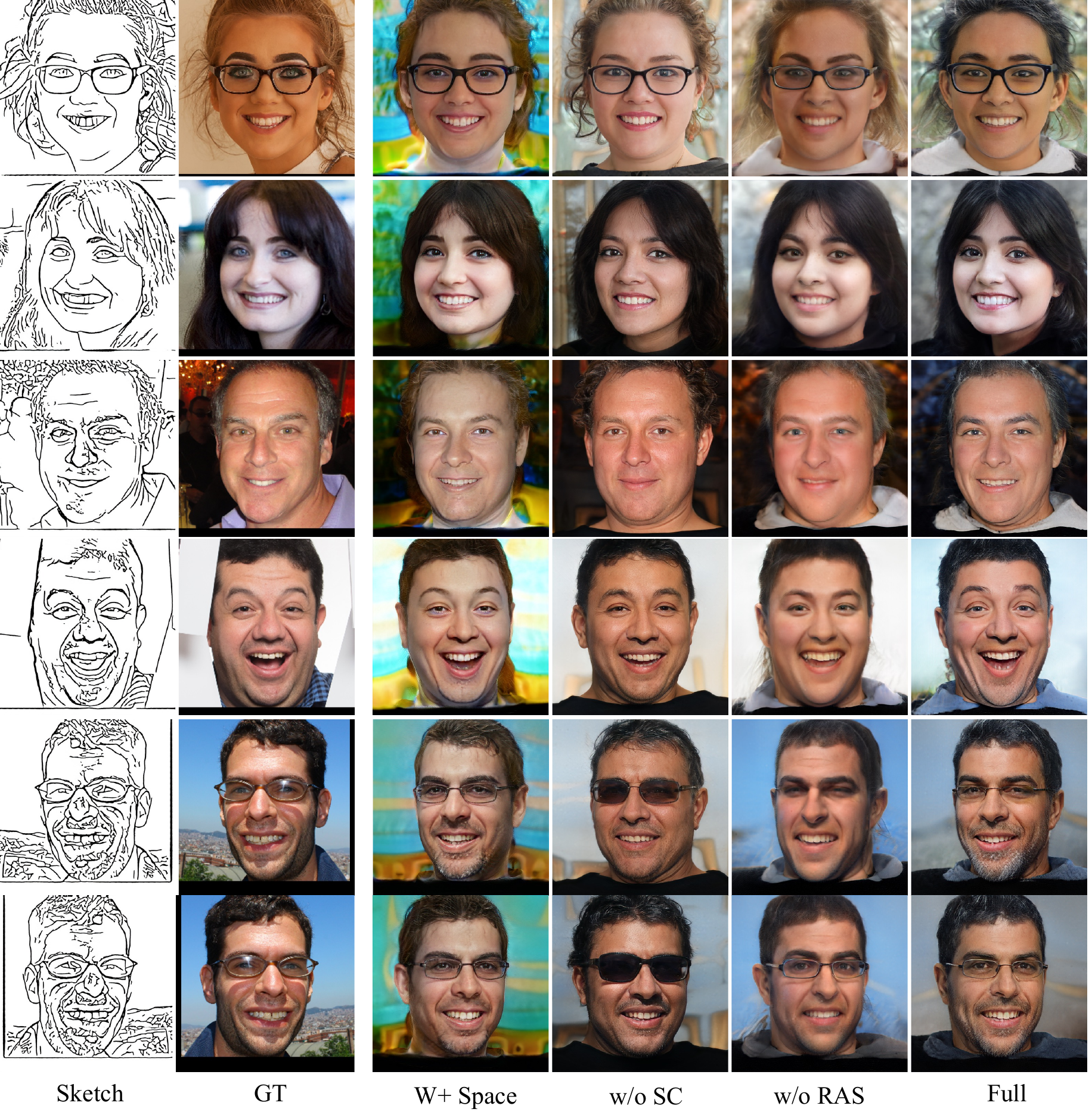}  
\end{center}
\vspace{-5pt}

\caption{\textbf{Qualitative Comparison of Ablation Study.} The visualization of different varieties of our method.
} 
\vspace{-10pt}
\label{ablation}
\end{figure}

\section{Usability Study}

To evaluate the usability of our SSSP system, we invite 15 participants (denoted as U1-U15) to conduct a usability study, composed of a fixed-task study and an open-ended study with a graphical interface we developed.

Among the users, 6 are females and 9 are males. Their ages are uniformly distributed from 20 to 40. Two of the users are product designers, three are computer science researchers, and two are artists. The others do not work on related topics.  We first ask all the participants to rate their drawing skills from 1 to 5 (rate 5 is the highest) and 80\% of them are amateur or middle users (scores 1 to 3). Then each participant requires to perform a fixed-task drawing session for learning how to use our graphical interface. After they warm up and are familiar with the system, an open-ended drawing session is conducted to create expected portraits without any limitations.

\begin{figure}
\begin{center}
\includegraphics[width=1.0\linewidth]{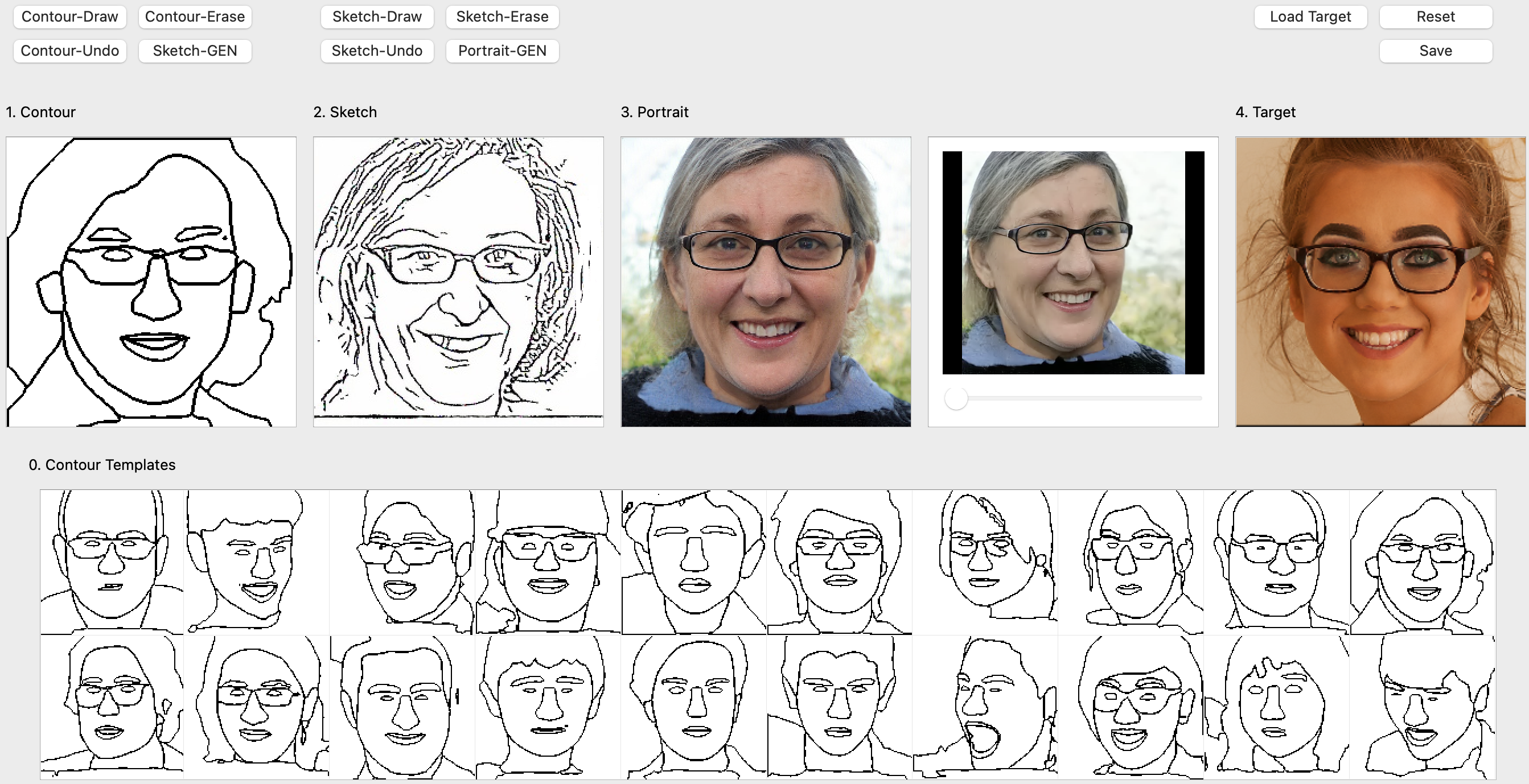}
\end{center}
\vspace{-5pt}
\caption{\textbf{Graphical Interface Design}.} 
\label{fig:interface}
\vspace{-15pt}
\end{figure}

\subsection{Graphical Interface Design}
The visualization of the interface is shown in Fig.~\ref{fig:interface}. It provides tools for fast editing and creating users' desired portraits from both coarse and fine sketches. 

\noindent\textbf{Contour Selection.} Though the users can directly create their sketches from scratch, drawing a reasonable contour has already been proven difficult for most amateur users. We provide a collection of portrait contours with different genders, hairstyles, facial shapes, and poses. The users can select one of the given contours as an easy start of this system, which saves quite a lot of effort.

\noindent\textbf{Contour Editing.} We provide contour \emph{drawing}, \emph{erasing} and \emph{undo} functions in the contour editing stage like other common image editing applications. The users could create or edit the coarse contour at this step. After they finish the contour creation, they can click on the ``Sketch-Gen'' button. The drawn contour will be sent into our Contour-to-Sketch module to synthesize a detailed sketch. The whole generation process takes less than 1 second and thus does not affect the users' experiences.

\noindent\textbf{Sketch Editing.} The initial sketches here are created by the contours from the previous step. Users can still interact with the sketches in the same way as the contour editing. They can erase the unsatisfactory lines created by the networks and add their expected details. After clicking the ``Portrait-Gen', the sketch is fixed and sent into our Sketch-Guided Stereoscopic Portrait Generation module to guide the generation of a high-quality 3D-aware portrait. This process takes less than 2 seconds.



\subsection{Fixed-task Study}
In this session, we provide several real portraits as target images for helping the users learn our system and practice their drawing skills. Each participant is requested to fill in a questionnaire and rate from 1 to 5 (1 for strongly disagree, 5 for strongly agree) on the following six aspects:
\begin{itemize}
    \item [1)] if our system is easy to use;
    \item [2)] if our system is compactly designed;
    \item [3)] if our system generates results consistent with your intentions;
    \item [4)] if our system offers sufficient guidance;
    \item [5)] if the coarse-to-fine process eases your creation;
    \item [6)] if the 3D demonstration increases your motivation.
\end{itemize}
The rating results are shown in the upper part of Fig.~\ref{fig:user_rate}, and we can observe that over 60\% of the participants rate 4 (agree) or 5 (strongly agree) in all the six aspects, especially all the participants reach an agreement that the 3D demonstration truly increases their motivation for creating. U1 who is an amateur comments that ``The generated 3D portraits are really impressive, and I can somehow quickly find out where and how to edit after a few times 
trying. The 3D demonstration can easily show me which part I should pay attention to when drawing contours. This improves not only my drawing skills, but also my imagination for the 3D space''.

Most participants are satisfied with the design of our system and give high scores on ``Easy Use'' and ``Compactly Designed''. U5 who is a product designer comments that ``I feel the system is user-friendly and well-designed. The 3D effect of the results is really impressive. It will indeed attract users after perfection.''. Only very few (\textit{i.e.}, drawing skill 4 or 5 scores) rate 1 or 2 on ``Coarse-to-Fine'' and ``Sufficient Guidance'', as their explanation is they have already raised their own drawing habits and styles, they prefer creating works freely all by themselves without any guidance.
According to the results, our system is able to provide sufficient guidance and assistance which can effectively help amateur or middle users practice their drawing skills and create their expected portraits.

\begin{figure}[t]
\begin{center}
\includegraphics[width=1.0\linewidth]{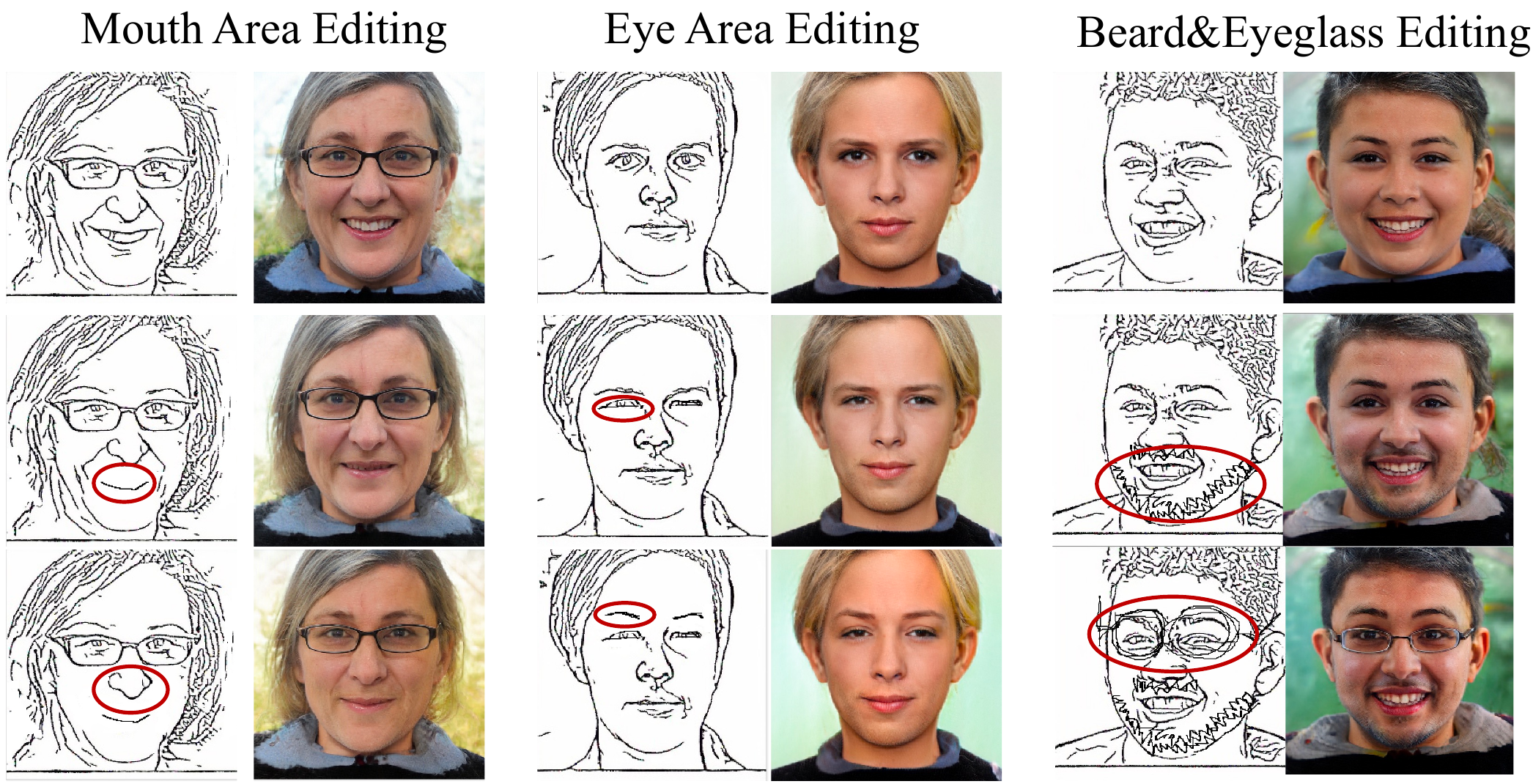}  
\end{center}
\vspace{-5pt}
\caption{\textbf{Samples of User Editing Operations.}
} 
\vspace{-15pt}
\label{local_edit}
\end{figure}

\subsection{Open-ended Study}
\begin{figure}[t]
\begin{center}
\includegraphics[width=1.0\linewidth]{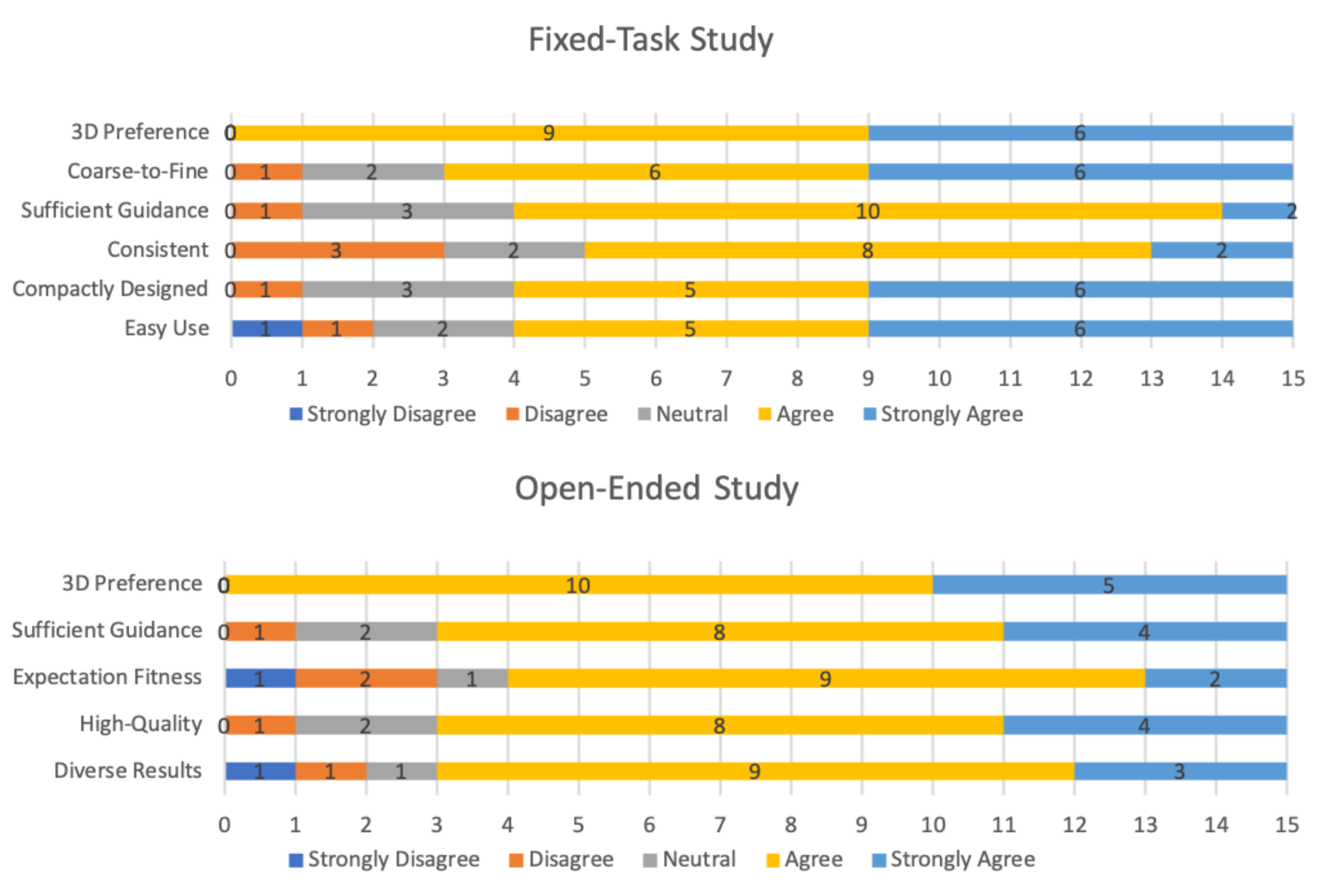}  
\end{center}
\vspace{-10pt}
\caption{\textbf{User Ratings on Fixed-Task  and Open-Ended Study.} 
} 
\vspace{-15pt}
\label{fig:user_rate}
\end{figure}

In the open-ended study session, the participants are requested to freely create their expected portraits with their hand-drawn sketches. Similarly, they are asked to fill in a questionnaire and rate from 1 to 5 (1 for strongly disagree, 5 for strongly agree) on five aspects:
\begin{itemize}
    \item [1)] if our system generates diverse results;
    \item [2)] if our system generates high-quality results;
    \item [3)] if our system fits your expectation;
    \item [4)] if our system supplies sufficient guidance;
    \item [5)] if the 3D demonstration increases your motivation.
\end{itemize}
The rating results are shown in the lower part of Fig.~\ref{fig:user_rate}. More than 73\% of the participants rate 4 (agree) or 5 (strongly agree) in all five aspects, among which all the participants agree or strongly agree that our 3D-aware portrait synthesis is impressive and effective when creating portraits. U7 comments that ``The 3D generation effect is reasonable, just exactly as people imagined''. U10 who is a content creation researcher comments that ``The quality of the generated images under multiple views are quite amazing. It would be very convenient for the artists to create 3D virtual human or other kind of 3D assets if this technique could be generalized to other categories.''

12 out of 15 participants agree or strongly agree with ``High Quality'' and ``Diverse Results'', which indicates that our proposed SSSP achieves robust performance even when the input free-hand sketches come from amateur users thanks to the delicately designed network architecture and training strategy.

Though we have shown the advantages of our SSSP system above, some participants also kindly leave constructive advice to us for improving our interactive system.
There are barely participants rate 1 or 2 on ``Sufficient Guidance'', ``High Quality'' and ``Diverse Results''
Few participants with better drawing skills do not agree with the ``Expectation Fitness'' aspect. U9 points that ``More editing functions should be developed, such as \textit{skin color}, \textit{hair color} and \textit{curly/straight hair}''. U12 comments that ``Common functions in the drawing software, such as zoom in/out and various line widths should be integrated into the system''. We will make our efforts to incorporate those functions commonly used in the commercial software into our interactive system to make it more user-friendly.

\section{Conclusion}

In this paper, we propose the Stereoscopic Simplified
Sketch-to-Portrait (SSSP) system, which generates high-quality stroke-consistent portraits with 3D awareness. Compared with previous methods, our system enjoys several intriguing properties. \textbf{1)} Our system directly converts sketches to portraits with 3D representations, thus we can freely render its views from the 3D space. This makes our system inherently different from previous studies. \textbf{2)} We propose the Contour-to-Sketch module which robustly converts sparse contours to detailed sketches. Such a topic has rarely been discussed. \textbf{3)} We build a user-friendly graphical interface where the users can freely choose to edit simple contours or sketches. Extensive user studies show that our system is greatly preferred by users.

\bibliographystyle{ACM-Reference-Format}
\bibliography{sample-base}
\appendix

\end{document}